\documentclass[letterpaper, 10 pt, conference]{ieeeconf}  

\IEEEoverridecommandlockouts                              

\usepackage{algorithm}
\usepackage{algpseudocode}
\usepackage{listings}
\usepackage{color}
\usepackage{xcolor,colortbl}

\usepackage{tabularx}
\usepackage{cellspace}
\usepackage[Export]{adjustbox}
\usepackage{subfig}
\usepackage{lipsum}
\usepackage{amsfonts}
\usepackage{subcaption}
\usepackage{hyperref}
\usepackage{tikz}
\usepackage{pgfplots}
\usepackage{graphicx}
\usepackage{cleveref}
\usepackage{caption}
\newsavebox{\arrangebox}

\title{\LARGE \bf
SceneSense: Diffusion Models for 3D Occupancy Synthesis from Partial Observation
}

\author{Alec Reed  \ \ \ Brendan Crowe \ \ \ Doncey Albin \ \ \ Lorin Achey \ \ \ Bradley Hayes \ \ \ Christoffer Heckman
\thanks{*This work was supported by NSF Award \#1932189.}
\thanks{All authors are with the Intelligent Robotics Laboratory, Department of Computer Science, at the University of Colorado Boulder,
        {\tt\small firstname.lastname@colorado.edu}}
}

\begin{document}

\maketitle
\thispagestyle{empty}
\pagestyle{empty}

\begin{abstract}
 When exploring new areas, robotic systems generally exclusively plan and execute controls over geometry that has been directly measured. When entering space that was previously obstructed from view such as turning corners in hallways or entering new rooms, robots often pause to plan over the newly observed space. To address this we present SceneScene, a real-time 3D diffusion model for synthesizing 3D occupancy information from partial observations that effectively predicts these occluded or out of view geometries for use in future planning and control frameworks. SceneSense uses a running occupancy map and a single RGB-D camera to generate predicted geometry around the platform at runtime, even when the geometry is occluded or out of view. Our architecture ensures that SceneSense never overwrites observed free or occupied space. By preserving the integrity of the observed map, SceneSense mitigates the risk of corrupting the observed space with generative predictions. While SceneSense is shown to operate well using a single RGB-D camera, the framework is flexible enough to extend to additional modalities. SceneSense operates as part of any system that generates a running occupancy map `out of the box', removing conditioning from the framework. Alternatively, for maximum performance in new modalities, the perception backbone can be replaced and the model retrained for inference in new applications. Unlike existing models that necessitate multiple views and offline scene synthesis, or are focused on filling gaps in observed data, our findings demonstrate that SceneSense is an effective approach to estimating unobserved local occupancy information at runtime. Local occupancy predictions from SceneSense are shown to better represent the ground truth occupancy distribution during the test exploration trajectories than the running occupancy map. Finally, we analyze example predictions and show that SceneSense provides reasonable, accurate, and useful predictions.
\end{abstract}    
\section{Introduction}
\label{sec:intro}

\begin{figure}[t!]
\vspace{5pt}
    \centering
    \includegraphics[width=200pt]{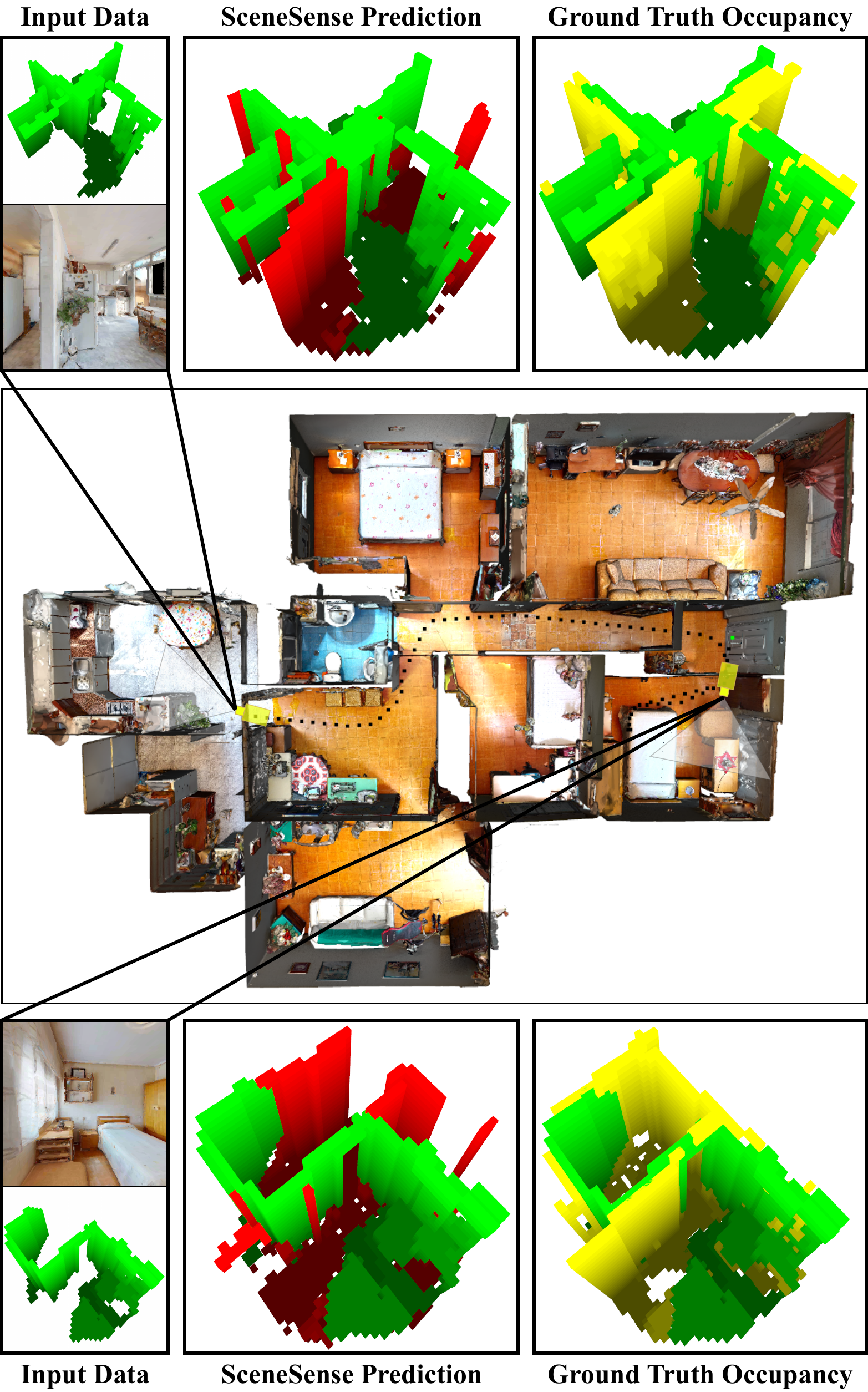}
    \caption{Test house 2 where the robot exploration trajectory is shown via the black points, and the starting point is shown as green. Two SceneSense generations are shown. From left to right (1) Inputs are on the left where green voxels are the local occupancy information as well as the current camera view from the robot. (2) SceneSense occupancy prediction is shown where occupancy information is shown in green and new predicted occupancy is red. (3) The running occupancy information is again shown in green and the ground truth full local occupancy data is shown in yellow.}
    \label{fig:h2_frames}
\vspace{-7pt}
\end{figure}
Humans rely extensively on `common sense' inferences to engage successfully with the world, while robots are limited to making decisions over directly measured data, such as those captured by cameras or lidar. Humans' natural capability to logically extend geometry or terrain in familiar environments such as homes or offices allows for planning beyond direct observation. In this work, we propose a solution addressing this important technical gap, to expand the scope of situations where autonomy can succeed.

Recent advances in AI systems that generate open-ended representations
, known as generative AI, give us the building blocks to develop a generative model for predicting out of view or occluded geometry.  Previous attempts at generating ``extended terrain'' borrowing from point cloud completion (PCC) methods \cite{reed2023looking} struggle to generalize to new environments. Existing methods for generating out-of-view geometry such as semantic scene completion (SSC) \cite{cheng21a_s3cnet,song2016semantic,yang2023diffusion} and more recently scene synthesis based approaches \cite{gao2022nerf, tang2023diffuscene} are promising. However, SSC is limited as it is a completion or hole-filling method applied only to the frustum of the sensor, rather than a truly generative approach that allows for full, 360$^\circ$ occupancy prediction.
For their part, synthesis-based approaches require many views for scene synthesis and are not usable as an online method due to slow inference speed. 

Motivation for development of these generative models can be found in the results of the DARPA subterranean (SubT) challenge \cite{chung2023into}. The SubT challenge tasked robot teams with the goal of locating human artifacts after being released into various unknown environments. These search and rescue missions provide a challenging and impactful venue for the deployment of autonomous systems. While teams were fairly successful in locating artifacts, the search took place with a long tail of artifact discovery over an hour. Systems over-searched areas to ensure maximum volumetric frontier gain and frequently paused when there was an influx of new information (such as turning corners into hallways or entering new rooms) \cite{Biggie2023Marble}. It is our hypothesis that platform exploration speeds can be accelerated using generative models for occupancy prediction.

In this work we leverage recent advances in generative AI as well as practically available robotics data to generate a predicted occupancy grid around a robotic platform. We show that even with a single RGB-D camera and limited training data we can develop an occupancy prediction model that effectively infers the existence of geometry that is out-of-view or occluded. During training, Gaussian noise controlled by a noise scheduler \cite{ho2020denoising} is added to ground truth occupancy data to generate noisy occupancy grids. We simultaneously train a U-net and perception backbone to reduce noise in the occupancy grid, conditioned by an RGB-D image. At inference time, features are extracted from the input images using the trained PointNet++ model \cite{qi2017pointnet} and used as conditioning during the reverse diffusion process. Additionally, we take advantage of the occupancy map constructed during exploration to perform \emph{occupancy inpainting}, increasing the fidelity of our results. Critically, using an inpainting approach ensures that the predicted occupancy grid around the robotic platform will never be modified in areas of the scene that have been directly observed to be occupied or free. Finally, we evaluate our framework in home environments from the HM3D dataset \cite{ramakrishnan2021hm3d} against a running occupancy map. Our results show that our proposed approach (\emph{SceneSense}) enhances the local occupancy predictions around the platform.  The primary contributions of this work are as follows:
\begin{enumerate}
    \item A generative framework for estimating out of view or occluded occupancy around the robotic platform.
    \item A diffusion inference method we call \emph{occupancy inpainting} that both enhances the predictions of the generative framework and ensures predictions will never overwrite observed free or occupied space. 
    \item An extensive ablation study outlining the performance trade offs of various tunable parameters that are configurable at runtime. 
\end{enumerate}

\section{Related Works}
\label{sec:RW} 
\subsection{Semantic Scene Completion (SSC)} 
SSC seeks to generate a dense semantically labeled scene in a target area given some sparse scene representation in that area. Generally the provided information is quite sparse and requires the SSC models to fill in large gaps due to occlusions from the viewpoint. SSC methods are often designed specifically for outdoors \cite{cheng21a_s3cnet} or indoor applications \cite{chen20203dsketch}. While outdoor SSC implementations focus on SSC using a 3D lidar, indoor methods use aimed sensors such as a RGB-D camera. Due to the shape of this input data outdoor models focus on prediction and labeling all voxels in a grid around the platform, while indoor models generally focus on performing SSC in the frustum of the sensor. Predicting correct geometry and semantic labels is a challenging task. SSC methods have been noted to suffer from poor performance given the challenging nature of the problem \cite{roldao20213sscSurvey}. Indoor methods perform around $40\%$ mIoU \cite{roldao20213sscSurvey} when ``filling in" data in the frustum of the sensor, but are not intended to be generative models that can expand a partially observed scene. Our method expands the role of these models to not only fill in occluded information in the sensor's field of view but to also generate a prediction of what geometry may look like around the platform. 
\subsection{Generative 3D Scene Synthesis}
View synthesis is a field of study that seeks to construct a 3D scene from multiple camera views. This is primarily accomplished using a deep learning framework called Neural Radiance Fields (NeRFs) \cite{gao2022nerf}. Original works in this space synthesized 3D views of objects from multiple camera views \cite{mildenhall2020nerf} , while recent works have synthesized full indoor scenes \cite{roessle2022dense}. Current NeRF implementations are slow at inference time, often taking on the order of minutes to render a scene \cite{gao2022nerf}. This characteristic makes them unsuited to real-time scene rendering. In addition, NeRFs require multiple views of the environment to generate reliable volumetric scenes, which may not be available when operating in real-time. 

\begin{figure*}[t!]
\vspace{3pt}
\centering
\includegraphics[width=\textwidth, height= 210pt]{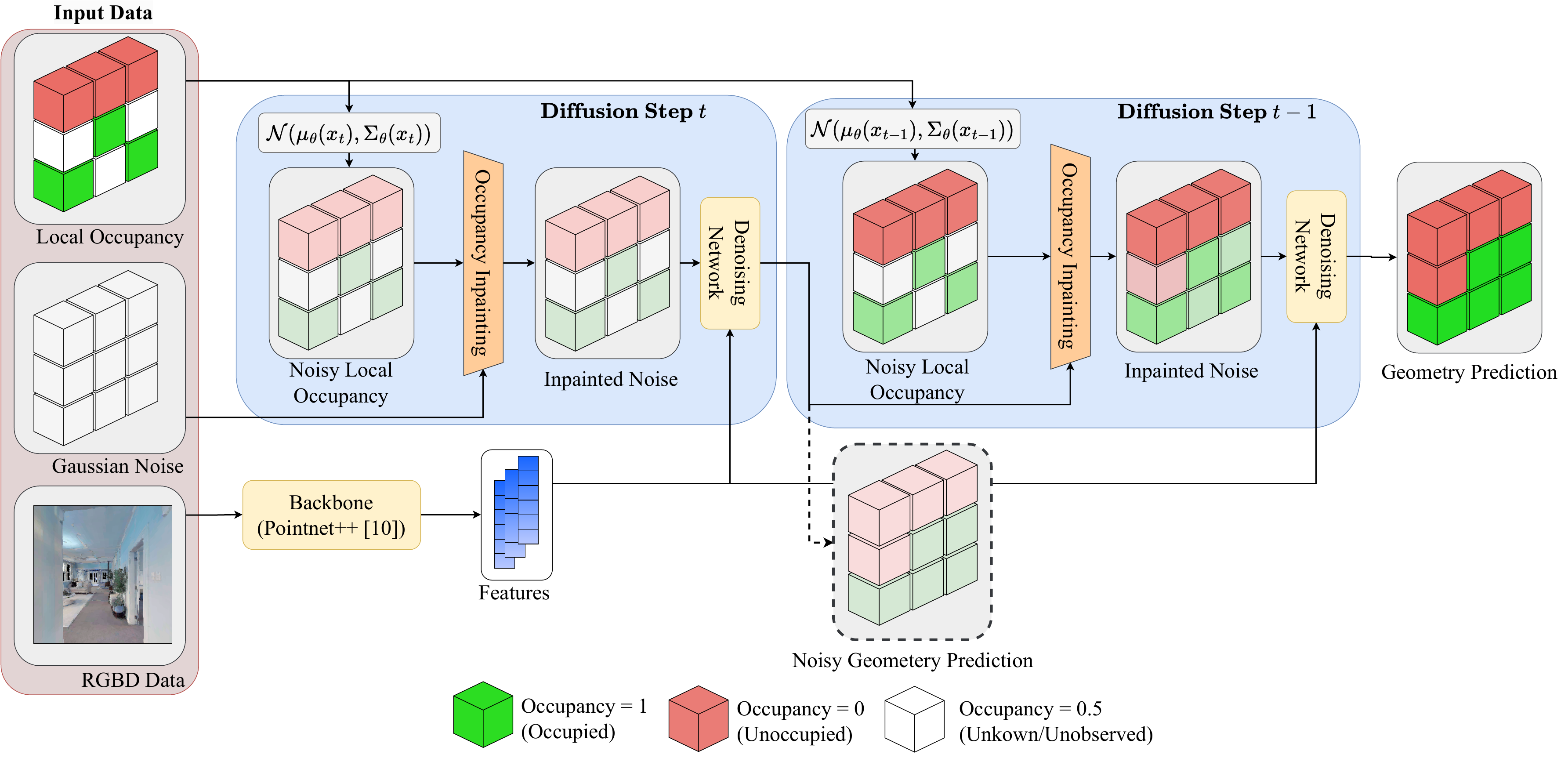}
\caption{\textbf{Reverse Diffusion Process}: The reverse diffusion process takes the local occupancy information, the current sensor measurements (RGB-D image in this case) and the Gaussian noise of the area to be diffused over. Noise commensurate with the current diffusion step is added to the local occupancy information, which includes occupied (green) and observed unoccupied (red) data. The result is inpainted into the noisy local occupancy prediction as discussed in \cref{sec:method}. The inpainted noise data and the feature vectors generated by the perception backbone are provided to the denoising network which generates a new noisy geometry prediction at $t-1$.   This processes is repeated as the starting noise $x_T$ is iteratively denoised to $x_0$ which is the final geometry prediction from the framework.}
\label{fig:Diffusion fwd}
\end{figure*}
Recently, diffusion models have been explored as a means for synthesising scenes \cite{tang2023diffuscene}. Initial studies show these models outperform traditional models in scene synthesis and introduce popular generative metrics to the scene synthesis research space. While these implementations are not directly usable in a robotics context, the ideas and evaluation metrics for 3D scene synthesis are applicable to our problem space. 
\subsection{3D Diffusion and Diffusion in Robotics}
Diffusion models \cite{ho2020denoising,pmlr-v37-sohl-dickstein15} have had great success as deep generative models. Diffusion models have generated impressive results across diverse modalities, such as image \cite{rombach2022highresolution} video \cite{harvey2022flexible}, audio \cite{huang2022prodiff} and natural language \cite{austin2023structured}. A number of surveys have been published in recent years providing further details on various implementations \cite{yang2023diffusion, croitoru2023diffusion}.  Research on diffusion in 3D is limited and generally applied to single object generation \cite{zeng2022lion}. However recent works have begun to explore application of diffusion models for scene generation. Notable work in this space include LegoNet \cite{yu2022legonet} which applies diffusion models to rearrange objects in a 3D scene and DiffuScene \cite{tang2023diffuscene} which supports unconditional or prompted diffusion of 3D scenes. While these diffusion methods are effective at their task they do not directly translate to robotic applications due to inference time requirements and the required conditioning data. 
\vspace{5pt}\\
\textbf{Diffusion for Robotic Applications}. \ \ 
The success of diffusion models have inspired researchers to begin to apply them in the robotic domain. While the Markovian nature of diffusion models can be a bottleneck for systems requiring real-time inference diffusion models have been successfully applied to real-time robotics problems such as  planning problems \cite{janner2022planning_w_diff, janner2022diffuser} and  perception \cite{ji2023ddp, chen2022diffusiondet}. These generative models are both increasing the effectiveness of current methods in traditional robotics problems as well as providing new research areas to tackle.
\section{Preliminaries and Problem Definition}

\label{sec:prelims}
\subsection{Problem Definition: Dense Occupancy Prediction}
The objective of dense occupancy prediction is to predict the occupancy from $[0,1]$ where 0 is unoccupied and 1 is occupied for every voxel $v$ in a target region $x$ where $v \in \mathbb{R} ^{\textbf{z}\times \textbf{x}\times \textbf{y}}$. 
\subsection{Forward Diffusion}
$x_0$ is defined as a clean occupancy grid where the distribution of $x_0$ can be defined as $q(x_0)$. By sampling from the data distribution $x_0 \sim q(x_0)$ the forward diffusion process is defined as a Markov chain of variables $x_1, ...,  x_T$ that iteratively adds Gaussian noise to the sample. A diffusion step at time $t$ in this chain is defined as:
\begin{equation}\label{eq:mark_step}
    q(x_t|x_{t-1}) = \mathcal{N}(x_t;\sqrt{1- \beta_t}x_{t-1},\ \beta_tI),
\end{equation}
where t is the time step $t \in [1,T]$, $\beta_t$ is the variance schedule $0 \leq \beta_t \leq 1$ and $I$ is the identity matrix. The joint distribution of the full diffusion process is then the product of the diffusion step defined in \cref{eq:mark_step}:
\begin{equation}
    q(x_{1:T}|x_{0}) = \prod_{t=1}^{T} q(x_t|x_{t-1}).
\end{equation}
Conveniently we can apply the reparameterization trick to directly sample $x_t$ given $x_0$ using the conditional distribution:
\begin{equation} \label{eq:add_noise}
    q(x_t|x_{0}) = \mathcal{N}(x_t;\sqrt{\overline{\alpha}_t}x_{0}, (1-\overline{\alpha}_t)\mathcal{I}),
\end{equation}
where $x_t = \sqrt{\overline{\alpha}_t}x_0 + \sqrt{1 - \overline{\alpha}_t}\epsilon$ where $\alpha_t := 1 - \beta_t$, $\overline{\alpha}_t := \prod^t_{r=1}\alpha_s$, and $\epsilon$ is the noise used to corrupt $x_t$.
\subsection{Reverse Diffusion}
Reverse diffusion is a Markov chain of learned Gaussian transitions $p_\theta (x_{t-1} | x_t)$ which is parameterized by a learnable network $\theta$:
\begin{equation}
    p_\theta(x_{t-1}|x_t) := \mathcal{N}(x_{t-1}; \mu_\theta (x_t , t) , \Sigma_\theta (x_t,t)),
\end{equation}
where $\mu_\theta (x_t , t)$ and $\Sigma_\theta (x_t,t)$ are the predicted mean and covariance respectively of the Gaussian $x_{t-1}$. 
Given the initial state of a noisy occupancy map from a standard multivariate Gaussian distribution $x_t \sim \mathcal{N}(0,I)$ the reverse diffusion process iteratively predicts $x_{t-1}$ at each time step $t$ until reaching the final state $x_0$ which is the goal occupancy map.  Similar to the Markov chain defined forward diffusion process the joint distribution on of the reverse diffusion process is simply the product of the applied learned Gaussian transitions $p_\theta (x_{t-1} | x_t)$:
\begin{equation}
    p_\theta(x_{0:T}) := p_\theta(X_T) \prod^T_{t=1}p_\theta(x_{t-1} | x_t).
\end{equation}
 \subsection{Conditional Diffusion}
 The conditional diffusion model extends the diffusion process to guide the diffusion by some conditioning $y$. In particular we use a class of conditional diffusion models called classifier-free diffusion \cite{ho2022classifier}. During training the diffusion model $f_\theta(x_t, y, t)$ is trained to predict $x_0$ from $x_t$ under the guidance of condition $y$. During training conditioning $y$ is replaced with a null label $\emptyset$ with a fixed probability. At inference time $x_0$ is reconstructed from $x_T$ with guidance from the conditioning $y$. During sampling at inference time the output of the model is ``pushed" toward the conditional model result $f_\theta(x_t |y)$ and away from the unconditioned result $f_\theta(x_t |\emptyset)$ as follows:
 \begin{equation} \label{eq:guidance_form}
     \hat{f}_\theta (x_t | y) = f_\theta(x_t | \emptyset) + s \cdot (f_\theta(x_t |y) - f_\theta(x_t | \emptyset)),
 \end{equation}
where $s$ is the guidance scale defined as $s \in \mathbb{R}_{\geq 0}$. $s$ is a configurable parameter at runtime that allows for the user to configure how closely the model should adhere to the provided conditioning. 

\section{Method}
\label{sec:method}
\subsection{Architecture}
\begin{figure}
    \vspace{7pt}
    \centering
    \includegraphics[width=180pt]{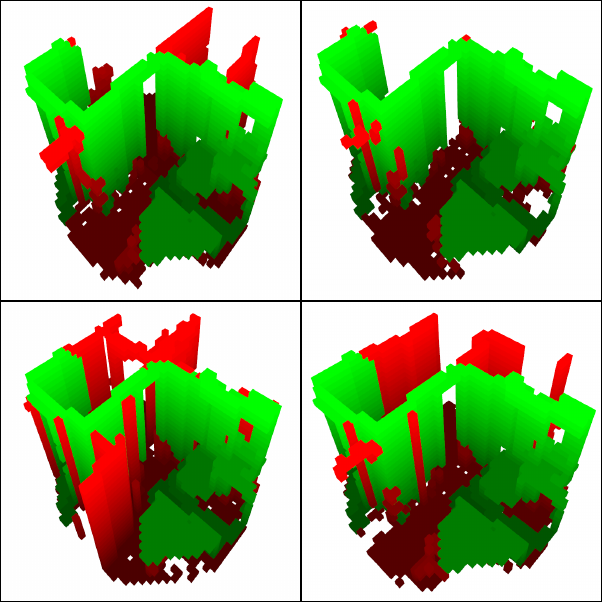}
    \caption{Various SceneSense predictions from equivalent input data where green is the running occupancy map and red is the SceneSense predicted occupancy. Given the limited input information the diffusion framework can generate multiple reasonable predictions from the same input conditioning.}
    \label{fig:same_cond_diff_gen}
\end{figure}
\noindent
\textbf{Denoising Network}. \ The denoising network in our method is inspired by the popular image generation diffusion network Stable Diffusion \cite{rombach2022highresolution}.  It is a U-net constructed from the HuggingFace Diffusers library of blocks \cite{von-platen-etal-2022-diffusers} and consists of Resnet \cite{he2015resnet} downsampling/upsampling blocks with cross-attention as well as regular ResNet downsampling/upsampling blocks. The conditioning features generated by Pointnet++ are mapped to the intermediate layers of the U-net via the cross attention layers of the transformer blocks as discussed by Stable Diffusion \cite{rombach2022highresolution}.
\vspace{5pt}\\
\textbf{Feature Extraction and Conditioning}. \ As discussed in \cref{sec:prelims} classifier-free diffusion models are conditioned on a set of guidance data $y$. Our model uses the Pointnet++ \cite{qi2017pointnet} backbone to generate a $N \times F$ feature matrix from a given RGB-D image, where $F$ is the number of desired Pointnet++ features per point $n \in N$. As is common with other vision based pipelines \cite{liu2022bevfusion} the feature generation backbone can be replaced with other models to increase performance or add/remove different types of perception modalities. Further, the conditioning of the diffusion model can be any modality encoding into feature vectors. Conditioning could include standard robotic sensors such as camera, lidar or radar, but could also be extended to include informational modalities such as human text input or sketches of the scene \cite{rombach2022highresolution}.
\vspace{5pt}\\
\textbf{Occupancy Mapping}. \ Occupancy mapping allows for platforms to build a running map of areas that have been measured to contain matter using onboard sensors like lidar or RGB-D cameras. For our framework we use the popular occupancy mapping framework Octomap \cite{hornung13octomap} to generate an occupancy map as the platform explores the environment. Importantly,  Octomap provides a probability of occupancy $o \in [0,1]$ for every voxel in the map that has been observed using pose ray casting. This means that as we explore we will be maintaining not only a map of occupied areas $M_o$, but also a map of areas that have been measured to not contain any data $M_u$. These maps will later be used to inform SceneSense where occupancy predictions should be made.
\subsection{Training}
During training, we generate a noisy local occupancy map $x_{t}$ where $t \in [1,T]$ from a ground truth local occupancy map $x$. We train the diffusion model $f_\theta$ to predict the noise applied to $x_t$ given the associated RGB-D conditioning $y$. 
\vspace{5pt}\\
\textbf{Occupancy Corruption}. \ To corrupt each ground truth local occupancy map $x$ to train the network we add Gaussian noise to $x$ to generate $x_t$. This corruption process is defined in \cref{eq:add_noise} where the intensity of the noise is controlled by $\alpha_t$ which is configured by a linear noise scheduler \cite{ho2020denoising}. 
\vspace{5pt}\\
\textbf{Loss Function}. \
The network $f_{\theta}$ is trained using the calculated $l_2$ loss between the denoised $x_t$ prediction and the associated ground truth data $x$. $l_2$ loss is a popular diffusion loss function, however other loss functions such as cross-entropy loss or mean squared error can be applied and have had some success in similar diffusion frameworks \cite{glide-nichol22a, rombach2022highresolution, chen2022diffusiondet, ji2023ddp}.
\label{sec:experiments}

\subsection{Inference}
\noindent
\textbf{Sampling Process} \ The trained noise prediction network $f_\theta$ takes isotropic Gaussian noise $\mathcal{N}(0, \mathcal{I})$ as the starting point $x_T$ to begin the reverse diffusion process. The noise is iteratively removed by using $f_\theta$ and the associated RGB-D features $y$ to compute $x_{t-1}$  The RGB-D features $y$ are applied as conditioning during the process with the cross-attention mechanism \cite{NIPS2017_3f5ee243}.
\vspace{5pt}\\ 
\textbf{Occupancy Inpainting}. \ Our method of occupancy inpainting ensures observed space is never overwritten with SceneSense predictions. Additionally occupancy inpainting enhances the predictions from the models seen in \cref{fig:ablations} (c). Inspired by image inpainting methods seen in image diffusion \cite{rombach2022highresolution} and guided image synthesis methods \cite{meng2022sdedit}, occupancy inpainting continuously applies the known occupancy information to the diffusion target during inference. To perform occupancy inpainting we sub-select a section of our occupancy map that we will perform diffusion over. In that area we take the occupied map $M_o$ and unoccupied map $M_u$ to generate occupied and unoccupied masks, $Ma_o$ and $Ma_u$ respectively. These masks will be applied to the target the diffusion to only allow occupied and unoccupied voxels to be modified during each masking step. Finally we add noise commensurate with the current diffusion step and update the diffusion target with the noisy occupied and unoccupied voxel predictions. This process is repeated for each inference step and can be seen in \cref{fig:Diffusion fwd}. This method both increases the fidelity of the scene predictions and ensures the diffusion model does not predict or modify geometry in space that has already been observed.
\vspace{5pt}\\
\textbf{Multiple Prediction}.
Diffusion is a noisy process that can generate different results given the same context. In image generation this is a desirable characteristic as the framework can generate different results given the same prompt, increasing the diversity of the generated data. As shown in \cref{fig:same_cond_diff_gen} SceneSense has the same behavior as these networks and can generate different reasonable predictions based on the same input information. Further there is no compute time increase  (assuming enough compute is available) as multiple predictions can be done in parallel. For simplicity we simply generate one prediction at each pose, but additional heuristics or voting schemes could be added to the system to score multiple outputs and select preferable predictions. 

\section{Experiments}
\label{sec:experiments}
\begin{figure}
\vspace{5pt}
    \centering
    \includegraphics[width=200pt]{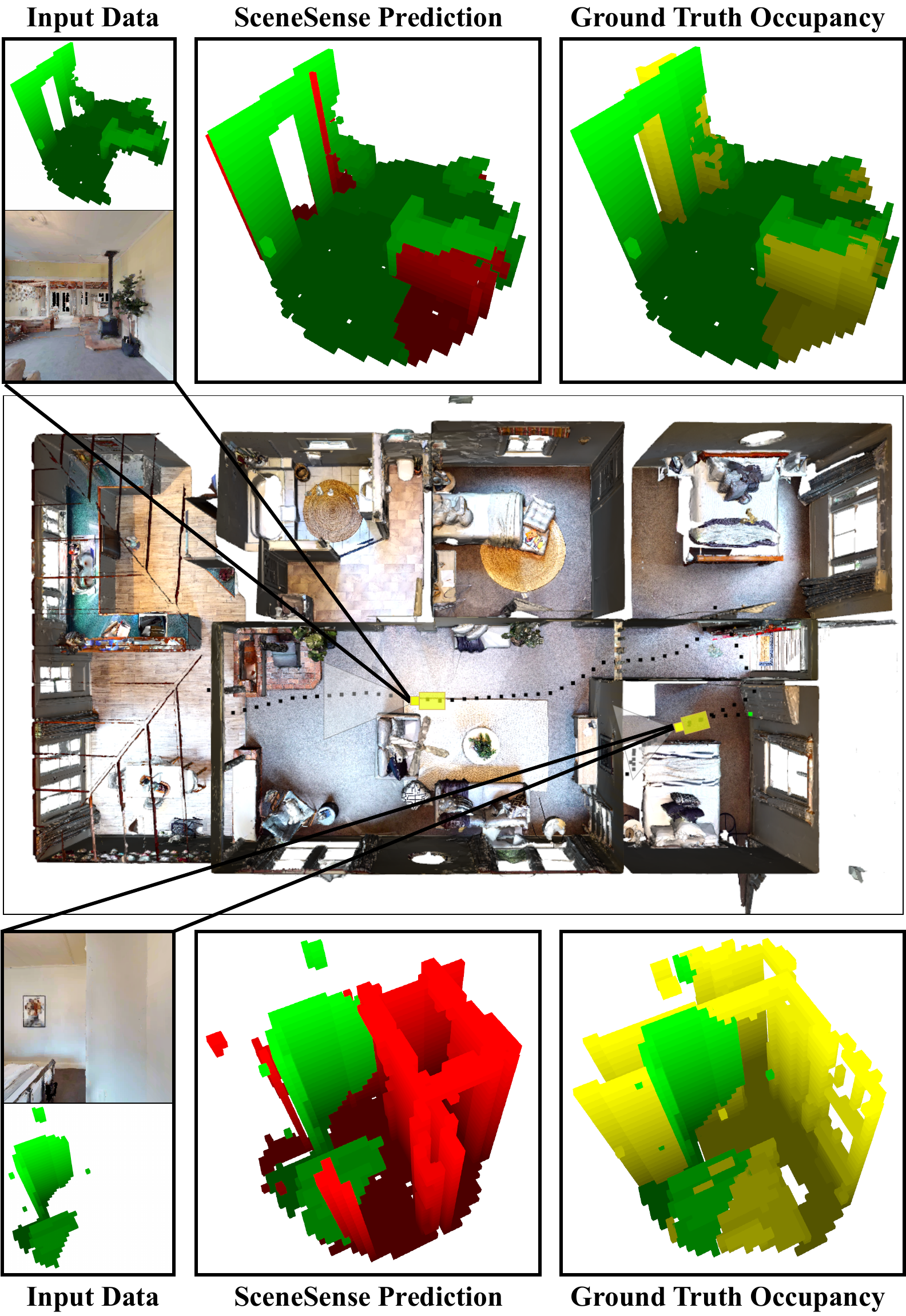}
    \caption{Test house 1 where the robot explore trajectory is shown via the black points, and the starting point is shown as green. Two SceneSense generations are shown. From left to right (1) Inputs are on the left where green voxels are the local occupancy information as well as the current camera view from the bot. (2) SceneSense occupancy prediction is shown where occupancy information is shown in green and new predicted occupancy is red. (3) The running occupancy information is again shown in green and the ground truth full local occupancy data is shown in yellow.}
    \label{fig:h1_frames}
\end{figure}
\begin{figure*}[t]
\centering
    \hfill\subfloat[][]{\includegraphics[width = 0.5\textwidth, height = 120pt]{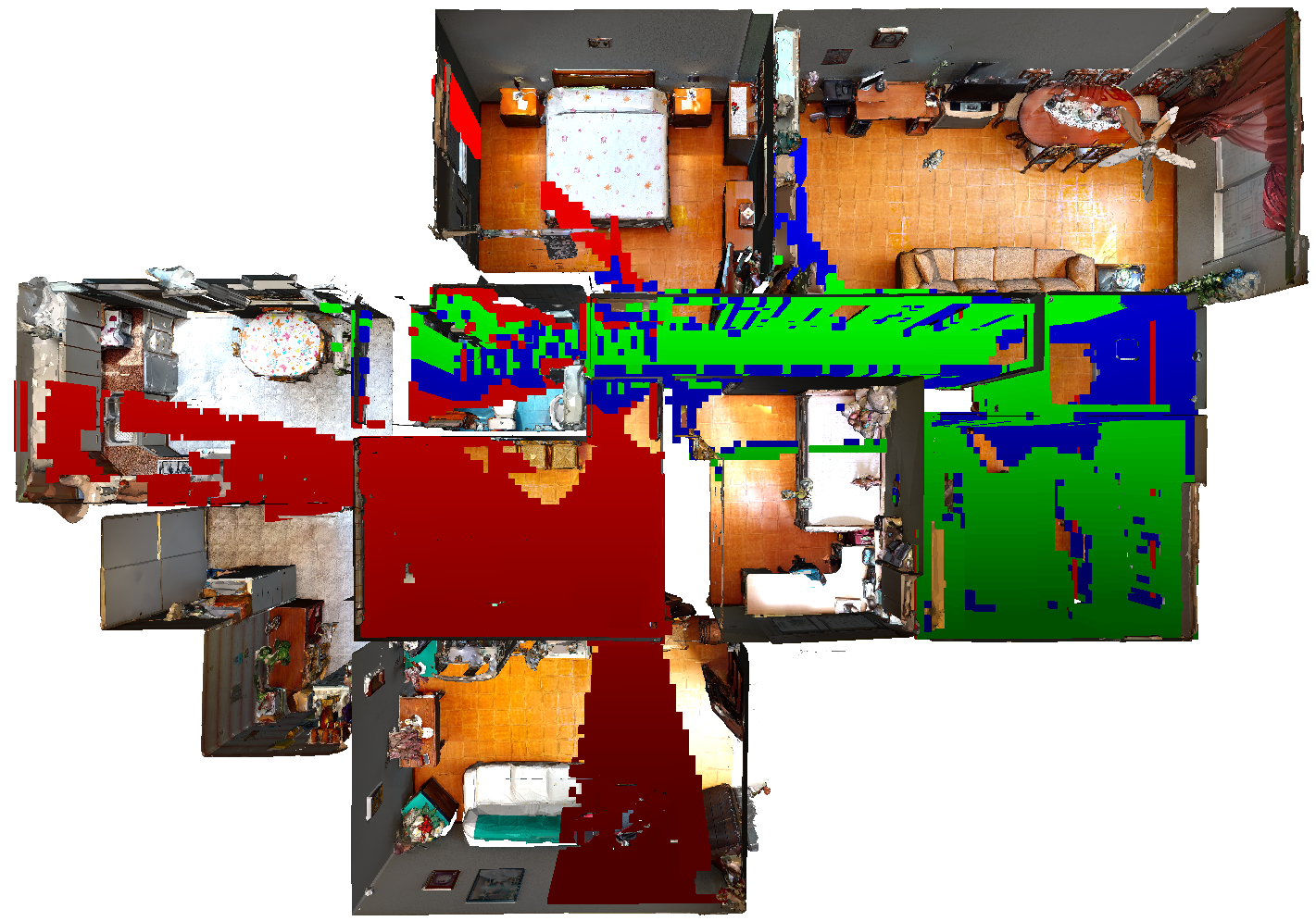}}\hfill
    \quad
     \subfloat[][\textbf{Conditioning}]{\resizebox{135pt}{!}{\includegraphics{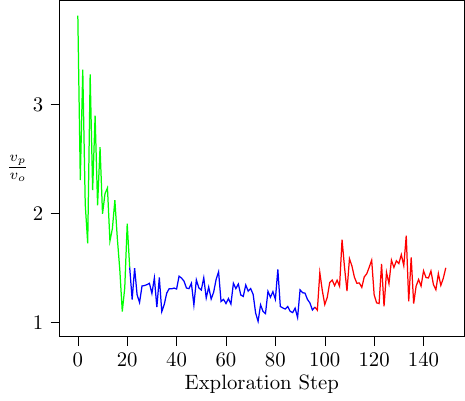}}}\hfill
    \quad
    \caption{Calculated predicted voxels $v_p$ over occupied voxels $v_o$ ($\frac{v_p}{v_o}$)  over the house 2 exploration. (a) Superimposes the running occupancy map over the house mesh where the colors of the occupancy map show how many steps have ran to that point. Green voxels are the running occupancy map from step 0 to step 20, blue are step 0 to step 95, and red are step 0 to 150. These colors correspond with the plot line colors in (b). (b) Shows the $\frac{v_p}{v_o}$ as the robot explores the space. $\frac{v_p}{v_o}$ starts high at time step 0, when the occupancy map is sparse, and quickly drops over the green exploration where more of the local scene is observed. $\frac{v_p}{v_o}$ stays relatively low as the vehicle completes the exploration of the green room, navigates back to the start point and traverses the hallway. $\frac{v_p}{v_o}$ increases slightly as the robot traverses previously unobserved space (red), which requests more predicted voxels as less of the scene has been observed.}
    \label{fig:vd_vo_plot}
\end{figure*}

We use the Habitat lab simulation platform \cite{puig2023habitat3} and the Habitat-Matterport 3D research dataset (HM3D) Dataset \cite{ramakrishnan2021hm3d} to generate training and test data. We operate a simulated platform with a 256$\times$256 RGB-D camera through 12 different house environment to generate full occupancy grids with voxel resolution of $0.1m$ of the homes as well as $\approx$9000 poses and associated RGB-D camera views to be used as conditioning. We split the dataset into a training and test set by house number. Houses 3--12 are used in the training set and houses 1 and 2 as shown in \cref{fig:h1_frames} and \cref{fig:h2_frames} respectively are used as the test set. For testing arbitrary trajectories are taken to navigate through each home, and SceneSense is used to predict the local occupancy information at each timestep. 
\vspace{5pt}\\
\textbf{Implementation}. \ 
 The diffusion model is trained using randomly shuffled pairs of conditioning $y$ and ground truth occupancy grids $x$, where various houses may be mixed in a batch. We use Chameleon cloud computing resources \cite{keahey2020chameleon} to train our model on one A100 with a batch size of 16 for 250 epochs or 119,250 training steps. We use a cosine learning rate scheduler with a 500 step warmup from $10^{-6}$ to $10^{-4}$. We set dropout to $0.2$ where the conditioning $y$ is set to $\emptyset$. The noise scheduler for diffusion is set to 1000 noise steps. At inference time we evaluate our dataset using an RTX 4090 GPU for acceleration. On average we measure a diffusion step for our model to be 0.0633 seconds.
\vspace{5pt}\\
\textbf{Baselines}. \ 
To our knowledge this is the first architecture to apply a generative method to predict 3D occupancy around a platform from a single aimed sensor. This makes direct comparisons for performance difficult to evaluate. As such, the best evaluation of our method is an evaluation against the running octomap (BL). Improvement upon this baseline shows that occupancy predictions from SceneSense better represents the ground truth occupancy information at a given pose than the running occupancy grid. 
The code and inference dataset will be released upon publication
\href{https://arpg.github.io/scenesense/}{https://arpg.github.io/scenesense/}.
\vspace{5pt}\\
\textbf{Evaluation}.  
Following similar generative scene synthesis approaches \cite{tang2023diffuscene, wang2021sceneformer}  we employ the Fr\'echet inception distance (FID) ] \cite{NIPS2017_8a1d6947} and the Kernel inception distance \cite{bińkowski2018demystifying} (KID $\times 1000$) to evaluate the generated local occupancy grids using the clean-fid library \cite{parmar2021cleanfid}. Generating good metrics to evaluate generative frameworks is a difficult task \cite{naeem2020reliable}. FID and KID have become the standard metric for many generative methods due to their ability to score both accuracy of predicted results, as well as diversity or coverage of the results when compared to a set of ground truth data. While these metrics are fairly new to robotics, which traditionally evaluates occupancy data with metrics like accuracy, precision and IoU, we show that they are an effective measure of the success of a generative framework like SceneSense.
\begin{table}[h]
\begin{center}
\caption{Quantitative comparisons of local occupancy synthesis from two test home environments from the HM3D \cite{ramakrishnan2021hm3d} dataset.}
\label{table:sim_results}
\begin{tabular}{l | c c | c c } 
\hline
 & \multicolumn{2}{c}{House 1} & \multicolumn{2}{c}{House 2}\\
Method & FID  $\downarrow$ & KID  $\downarrow$ & FID  $\downarrow$ & KID  $\downarrow$ \\
\hline
BL & 26.18 & 18.91 & 22.55 &  14.06\\ 
SS & 17.81 & 7.93 & 20.94 & 6.93\\
\hline
\end{tabular}
\end{center}
\end{table}
\section{Results} \label{sec:results}
Quantitative results for the test set are shown in \cref{table:sim_results} while qualitative results showing example diffused occupancy grids can be seen in \cref{fig:h1_frames} and \cref{fig:h2_frames}. SceneSense achieved substantial reductions in both KID and FID when compared to the baseline running occupancy method. Additionally, the qualitative results show reasonable estimates of potential geometry around the platform given limited input information.
\begin{figure*}[t]
\centering
    \hfill\subfloat[][\textbf{Denoising steps}]{\resizebox{150pt}{95pt}{\includegraphics{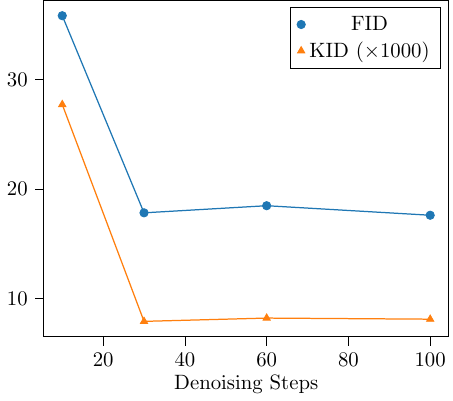}}}\hfill
    \quad
    \hfill\subfloat[][\textbf{Guidance scale $s$}]{\resizebox{150pt}{95pt}{\includegraphics{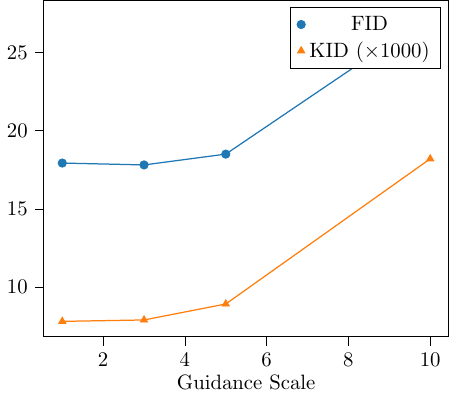}}}\hfill
    \quad
     \hfill\subfloat[][\textbf{Conditioning}]{\resizebox{150pt}{95pt}{\includegraphics{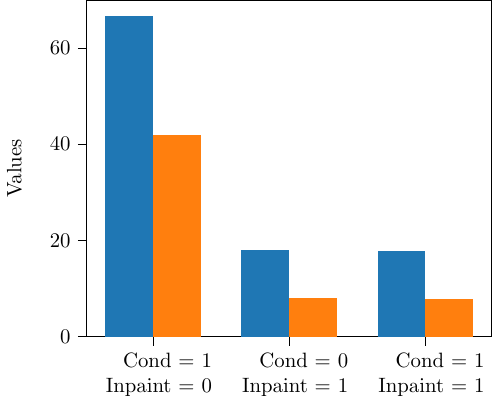}}}\hfill
    \quad
    \caption{\textbf{SceneSense ablation experiments}: All ablation experiments were run on test house 1 using the same trained diffusion network. For all experiments conditioning and inpainting are enabled, $s$ is set to 3 and 30 denoising steps are used unless these values are being ablated. (a) Figure (a) ablates various denoising step values. (b) Figure (b) ablates various guidance scale $s$ values as defined in \cref{eq:guidance_form}. (c) Figure (c) ablates enabling conditioning and inpainting for the network. A 1 indicates the value is enabled and 0 indicates it is disabled.}
    \label{fig:ablations}
\end{figure*}
\vspace{5pt}\\
\textbf{Quantitative Discussion}. \ The results presented in \cref{table:sim_results} favor SceneSense when compared to the local running occupancy information in both scenes tested. In house 1 substantial reduction in FID and KID, 31$\%$ and 58$\%$ respectively, were reported. While the reduction in FID was lower in house 2 (7$\%$) the reduction in KID was similar to that measured in house 1, 50$\%$. KID is known to be less sensitive to outliers and is considered by some to be an advantageous metric for evaluating generative frameworks when compared to the FID  \cite{bińkowski2018demystifying}. As shown in \cref{fig:h2_frames} house 2 is a much different layout than house 1, with many small hallways, rooms and corners. It is likely that house 1 is more similar to those captured in the training set than house 2, leading to an increase of erroneous predictions in the house 2 scene. These predictions result in a larger FID score while the KID remain low due to the native robustness to these skewed distributions.

Further we can look at example predictions for qualitative evaluation and to examine FID and KID as evaluation metrics compared to traditional metrics such as IoU. Take for example the upper prediction in \cref{fig:h2_frames} where the platform is entering the kitchen. SceneSense adds useful information to the problem, predicting the existence of floor as well as a wall that would obstruct motion to the right of the robot. However the wall is $0.1$ - $0.2$m off the actual existing wall. The result of this incorrect locating results in a worse IoU (0.52) than the local occupancy information (0.61). These problems in quantitative evaluation are only exacerbated when there is more geometry to estimate such as in the bottom observation of \cref{fig:h1_frames}. Generative models incur a large penalty in the IoU metric for guessing at geometry, even when given very little context for the prediction. Predictions increase the total union of the space, however a missed prediction does not increase the intersection of the space. In the case of the top prediction in \ref{fig:h2_frames} even if the prediction would be useful in planning, since the wall is mislocated the metric reports a much worse IoU than the running occupancy. Additionally in cases where little information is known, such as in the bottom prediction of \ref{fig:h1_frames}, an occupancy prediction of what the geometry may look like is heavily penalized since the metric does not evaluate if a prediction distribution is reasonable. However the FID and KID metrics evaluate both the accuracy of predictions as well as the distribution of the predictions, allowing for generative frameworks that attempt to generate challenging results to be fairly scored. FID and KID are already widely adopted metrics in generative fields such as image generation \cite{ramesh2022dalle, rombach2022highresolution} and scene synthesis \cite{tang2023diffuscene, wang2021sceneformer} for these reason and our results support their use in this context.
\vspace{5pt}\\
\textbf{SceneSense Accuracy Over Time}. As discussed in \cref{sec:method} SceneSense only predicts occupancy in areas that have not been directly measured to be occupied or unoccupied. This means that as exploration of the space approaches 100$\%$ SceneSense will have no unobserved space to predict over. This can be measured as $\frac{v_p}{v_o}$ where $v_p$ are the predicted voxels from SceneSense and $v_o$ are the local occupied voxels. Given 100$\%$ exploration this metric will approach 1 where all predicted voxels are simply the local occupied voxels. This reduction in prediction space is shown in \cref{fig:vd_vo_plot}. Initially $\frac{v_p}{v_o}$ is quite high, since very little of the scene has been observed but quickly drops during exploration. Spikes in $\frac{v_p}{v_o}$ are seen when the platform moves to new areas that have occlusions such as hallways or when entering new rooms. These metrics support the assertion that SceneSense respects measured space and only generates geometry where no measurement has been taken.

\subsection{Ablations}
\textbf{Denoising Steps Discussion}. \
The number of diffusion steps defines the size of each diffusion step during the reverse diffusion process. Generating reasonable results using the fewest possible denoising steps is desirable behavior to reduce computation time. Additionally too many denoising steps have been shown to introduce sampling drift which results in decreased performance \cite{chen2022diffusiondet, ji2023ddp}. As shown in  \cref{fig:ablations} (a) our method saw the best results when configured to 30 denoising steps. Too few denoising steps results in the network being unable make accurate predictions over the large time step. Increasing the number of denoising steps keeps results relatively stable over time, however you can see the KID is slightly worse using more steps due to sampling drift. Sampling drift is a result of the discrepancy between the distribution of the training and the inference data. During training, the model is trained to reduce a noisy map $x_t$ to a ground truth map $x$, at inference time the model iteratively removes noise from its already imperfect noise predictions. These predictions will drift away from the initial corruption distribution which becomes more pronounced at smaller time steps due to the compounding error.
\vspace{5pt}\\
\textbf{Conditioning and Guidance Scale Discussion}. \
The guidance scale $s$ as defined in \cref{eq:guidance_form} is a constant that multiplies the difference of the conditional diffusion and unconditional diffusion to ``push'' the diffusion process towards the conditioned answer. Setting $s$ too high results in too large of pushes away from reasonable predictions and results in poor generalization to new environments. The best FID is measured when $s = 3$ however KID is slightly lower when $s = 1$. Further, when examining chart (c) it is shown that the results when conditioning is removed all together ($s = 0$ ) are very similar to the best results seen with conditioning enabled (albeit slightly worse). This is likely because most of the useful conditioning information is captured in the local occupancy data, and mapping measured RGB-D points from area in front of the to geometry under or behind the platform is a very difficult task. The performance of the conditioning only data may be seen to have a larger impact on the overall results if the sensor could capture more local information, such as given a wider FOV or different mounting angle. 
\section{Conclusions and Future Work}
\label{sec:conclusion}
In this paper we present SceneSense; a diffusion-based approach for generative local occupancy prediction. SceneSense is shown to enhance local occupancy information quantitatively using standard metrics for generative AI, as well as qualitatively by providing frames generated by the framework. While past methods have focused on filling holes in observed information, or synthesising scenes offline from multiple camera views, SceneSense can quickly infer scene information 360$^{\circ}$ around the platform. Future work for SceneSense includes integration with a planning and control architectures for scene exploration and testing of additional conditioning modalities such as language to further enhance scene predictions.   

\bibliographystyle{unsrt}
\bibliography{main}

\end{document}